\documentclass[runningheads]{llncs}
\usepackage[T1]{fontenc} 
\usepackage[utf8]{inputenc} 
\usepackage[english]{babel} 
\usepackage{csquotes} 

\usepackage{xcolor}

\usepackage{amsmath, amssymb, mathtools, commath, bm} 
\usepackage{dsfont} 
\usepackage{bbm} 

\usepackage{graphicx} 
\usepackage{caption} 
\usepackage{subcaption} 
\usepackage{wrapfig} 
\usepackage{booktabs} 
\usepackage{hhline} 
\usepackage{array} 
\newcolumntype{C}{>{\centering\arraybackslash}X} 
\usepackage{adjustbox} 
\usepackage{svg} 

\usepackage{algorithm} 
\usepackage{algpseudocode} 
\usepackage{algorithmicx}
\floatname{algorithm}{Algorithm} 
\usepackage[vlined, linesnumbered, algochapter, algo2e]{algorithm2e}

\usepackage[inline]{enumitem} 

\usepackage{todonotes} 
\usepackage{mdframed} 
\usepackage{makecell} 
\usepackage{bbding} 
\usepackage{hyperref} 

\SetKwProg{Procedure}{}{:}{}
\SetKwIF{If}{ElseIf}{Else}{if}{:}{else if}{else :}{end if}
\SetKwFor{While}{while}{:}{end while}
\SetKwFor{Repeat}{repeat}{:}{end repeat}
\SetKwRepeat{Do}{do}{while}
\SetKwFor{ForEach}{foreach}{:}{end foreach}
\SetKw{KwDownTo}{downto}
\SetKw{KwTo}{to}

\usepackage[misc]{ifsym}

\usepackage{cite}

\graphicspath{{figures/}}

\newcommand{\etal}{\textit{et al}.}

\newcommand{\nk}[1]{\textcolor{purple!90}{#1}}

\usepackage{hyperref}
\usepackage{cleveref} 

\usepackage{ulem} 

\begin{document}

\title{LLMs in the Loop: Leveraging Large Language Model Annotations for Active Learning in Low-Resource Languages}

\author{Nataliia Kholodna \and Sahib Julka \thanks{Corresponding author:~\email{sahib.julka@uni-passau.de}}~\Letter \and Mohammad Khodadadi \and Muhammed Nurullah Gumus \and Michael Granitzer}

\institute{Chair of Data Science,
University of Passau, Germany}

\authorrunning{N. Kholodna \and S. Julka \and M. Khodadadi \and M. Gumus \and M. Granitzer}
\titlerunning{LLMs in the Loop}

\tocauthor{Nataliia Kholodna, Sahib Julka, Mohammad Khodadadi, Muhammed Nurullah Gumus, Michael Granitzer}
\toctitle{LLMs in the Loop: Leveraging Large Language Model Annotations for Active Learning in Low-Resource Languages}

\maketitle

\begin{abstract}
Low-resource languages face significant barriers in AI development due to limited linguistic resources and expertise for data labeling, rendering them rare and costly. The scarcity of data and the absence of preexisting tools exacerbate these challenges, especially since these languages may not be adequately represented in various NLP datasets. To address this gap, we propose leveraging the potential of LLMs in the active learning loop for data annotation. Initially, we conduct evaluations to assess inter-annotator agreement and consistency, facilitating the selection of a suitable LLM annotator. The chosen annotator is then integrated into a training loop for a classifier using an active learning paradigm, minimizing the amount of queried data required. Empirical evaluations, notably employing GPT-4-Turbo, demonstrate near-state-of-the-art performance with significantly reduced data requirements, as indicated by estimated potential cost savings of at least 42.45 times compared to human annotation. Our proposed solution shows promising potential to substantially reduce both the monetary and computational costs associated with automation in low-resource settings. By bridging the gap between low-resource languages and AI, this approach fosters broader inclusion and shows the potential to enable automation across diverse linguistic landscapes.

\end{abstract}

\section{Introduction}

Large Language Models (LLMs) like Claude 3 Opus~\cite{anthropic2023} and GPT-4~\cite{openai2023gpt4}, have demonstrated exceptional performance across various Natural Language Processing (NLP) tasks, particularly in high-resource languages such as English~\cite{Naveed2023A}. However, their effectiveness in low-resource languages remains underexplored due to limited comprehensive datasets for evaluation \cite{bommasani2022opportunities}. This research gap arises from both the scarcity of data and the high cost associated with human annotation, which is crucial for training and evaluating models in these languages~\cite{Neveol2011Semi-automatic, Ferragina2010Fast}.

Active learning (AL) strategies offer a promising solution by iteratively selecting informative samples to optimize resource utilization and accelerate model learning~\cite{Kholghi2017Active,Liao2016Visualization-Based}. These considerations motivate a key research question: Can the combination of state-of-the-art LLMs, effective prompting, and active learning strategies enable resource-efficient data annotation in low-resource languages? Although there have been some advances in tackling the difficulties of data annotation and model training in this context \cite{pangakis2023automated}, automating these processes through foundation models is still largely unexplored. 

With an overarching objective to minimize costs and resources, we introduce a novel methodology that integrates foundation models into an active learning framework to enhance the annotation process. This study investigates the application of our proposed methodology specifically in Named Entity Recognition (NER) tasks within low-resource language contexts, using African languages as a focal point for experimentation. The findings from this study hold potential for extending the methodology to other low-resource languages, thereby advancing progress in NLP across various domains. Concretely, our contributions can be listed as follows:
\begin{itemize}
\item We conduct a comparative evaluation of various popular LLMs (GPT-4-Turbo, Claude 3 Opus, Sonnet~\cite{anthropic2023}, Gemini 1.0~\cite{geminiteam2023gemini}, Mistral 7b~\cite{jiang2023mistral}, Llama 2 70b~\cite{touvron2023llama}, and Starling-LM 7b~\cite{starling2023}) for NER tasks in low-resource African languages, analyzing their performance differences in accuracy and consistency of annotation.
\item  We investigate common NER annotation challenges faced by LLMs, including token skipping and output parsing. We assess the impact of prompt quality and batch annotation on output accuracy.
\item We introduce a novel methodology to quantify data leakage by measuring potential data contamination.
\item We quantify the potential cost savings achieved by integrating LLMs into an active learning framework for data annotation, demonstrating significant reductions in cost and data requirements compared to traditional human-annotation methods\footnote{Open source code can be found at:~\url{https://github.com/mkandai/llms-in-the-loop}.}. 
\end{itemize}
\section{LLMs in the Loop}
\label{sec:method}

Large Language Models have rapidly advanced, with prompting methods like In-context learning (ICL) enabling LLMs to perform well in few-shot and zero-shot settings across various language tasks~\cite{brown2020language,yao2022react, kojima2022large, suzgun2022challenging, zhong2021adapting}. This includes information extraction~\cite{wei2023zero,wang2023gpt} and even achieving a performance that is often comparable to or exceeds that of human annotators for popular languages on crowdsourcing tasks~\cite{gilardi2023chatgpt, chiang2023can, ziems2024can, he2023annollm,bang2023multitask}. 

Named Entity Recognition (NER) is a sub-task of information extraction that focuses on identifying and classifying entities within text. These entities typically fall into predefined categories such as person names, organizations, locations, dates, etc.~\cite{ASurveyOnNER2007} Although preliminary work suggests that LLMs could act as active annotators for tasks such as NER~\cite{zhang2023llmaaa}, their performance in low-resource languages remains a critical research gap. 

We test the applicability of these models as ad-hoc annotators in an AL paradigm. This is aimed at satisfying two of our main objectives: a) reducing annotation costs, and b) saving computational resources.

Active learning is a machine learning technique that is based on selecting the most informative data points for training, enabling the task-specific model to learn more efficiently from a smaller dataset. By sampling data with high uncertainty in prediction, the aim is to maximize learning with a minimal amount of data. 
 
\begin{figure}[ht]
    \centering
    \includegraphics[width=0.7\linewidth]{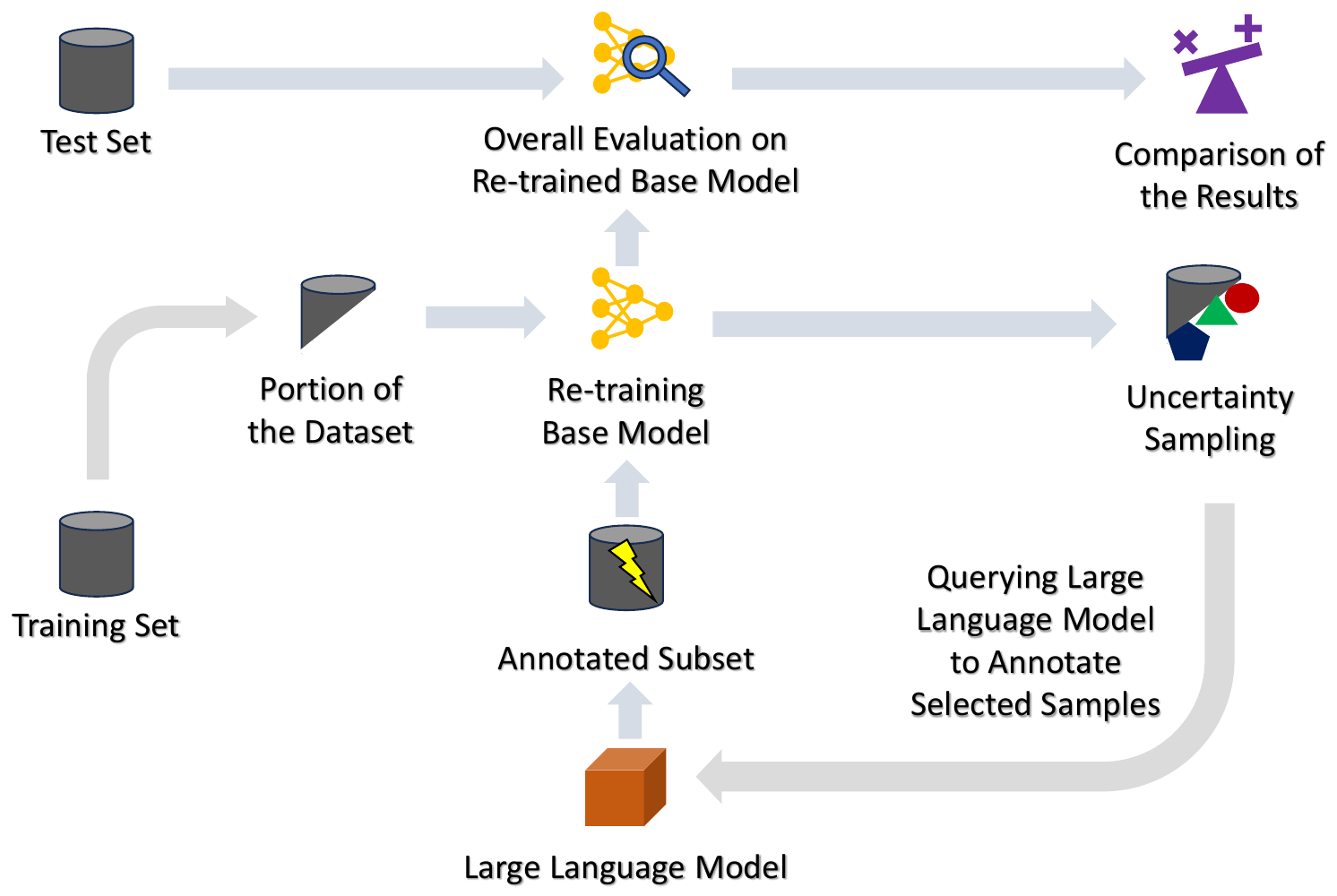}
    \caption{Overview of our methodology. The process involves selecting the most informative samples from the training set, and querying the LLM with a pre-defined prompt template to obtain annotations. The problem-specific classifier is then trained with these queried annotations and evaluated on the unseen test set.}
    \label{fig:methodology}
\end{figure}

We propose a novel approach for querying LLMs for NER labels using batching of samples~(cf.~\Cref{sec:querying_in_batches}). This results in the same prompt being used for annotation of multiple samples, effectively expediting the process and reducing the querying cost as fewer tokens are expended.
We adopt an iterative active learning approach~(cf.~\Cref{fig:methodology}) for fine-tuning a NER model on the MasakhaNER 2.0 dataset~\cite{Adelani2022MasakhaNER2A}.

MasakhaNER 2.0~\cite{Adelani2022MasakhaNER2A} is the largest human-annotated NER dataset for African languages that contains text data from 20 languages widely spoken in Sub-Saharan Africa. 
Despite these languages having a considerable number of speakers they are underrepresented in NLP research. 
The dataset contains a comprehensive collection of entities, including people, places, organizations, and dates. The percentage of annotated entities relative to all tokens ranges from 6.5~\% for Bambara to 16.2~\% for Shona language.

Our methodology consists of the following core steps:
\begin{enumerate}
    \item \textbf{Initial Model Training:} We fine-tune the AfroXLMR-mini model \cite{alabi-etal-2022-adapting} on a small subset (5~\%) of the MasakhaNER 2.0 dataset, establishing a baseline performance. AfroXLMR-mini is a compact multilingual language model that is based on the BERT~\cite{devlin2019bert} architecture, fine-tuned specifically for African languages.

    \item \textbf{Active Learning Loop:}  We iteratively execute the following:

        \textit{Uncertainty Quantification:}
            Using predictions from the model,  we calculate uncertainty at the token level using entropy: \begin{equation}H(i) = -\sum_{c=1}^{C} p(c|i) \log p(c|i),
                \end{equation}
                where for each token i, $p(c|i)$ is the probability of class c, with C  as the total number of entity classes.  We then compute the average sample entropy: 
                \begin{equation}H_{\text{avg}}(\text{sample}) = \frac{1}{|\text{sample}|}\sum_{i \in \text{sample}} H(i)\end{equation}

        \textit{Sample Selection:}  We rank samples by descending entropy and select the top 5~\% of the most uncertain samples for annotation with the best performing LLM of all compared. We add new annotations to the initial training subset and retrain the AfroXLMR-mini model on the expanded dataset.
\end{enumerate}

\section{Experiments}
\subsection{Foundation Model Selection}
\label{sec: foundation_model_selection}

Our experimental setup hinges on the initial choice of a suitable foundation model. In low-resource settings, where data are scarce and of high value, an accurate annotator is crucial for high-quality data labeling, particularly for NER. 

NER presents a significant challenge for LLMs because it demands not only accurate labeling of each token in the text without omissions but also requires the answers to be returned in a specific format, making it a more complex task than, for example, classification, where the model is tasked with providing a categorical value. 


We evaluated several LLMs as potential candidates, viz. GPT-4-Turbo [version "gpt-4-0125-preview"], Google Gemini 1.0 pro [version "gemini-1.0-pro-vision-001"], Llama 2 70B, Mistral 7B, and Starling-LM-7b-alpha. While Anthropic's Claude 3 Sonnet and Claude 3 Opus were assessed in a smaller-scale benchmark, their high operational costs precluded inclusion in our final analysis. 

Candidate selection is performed using the following criteria: a) adherence to the desired output format, b) correctness of the assigned labels, and c) labeling consistency and inter-annotator agreement~(IAA). We elaborate on them in the subsequent subsections.

\subsubsection{Representative Data Subset Sampling for LLM Evaluation}
\label{sec:sampling_approach}

To assess the performance of LLM annotation on the MasakhaNER 2.0 dataset, we conducted experiments on a stratified subset of its languages. To ensure the generalizability of our approach across languages with different resource levels and distinct linguistic characteristics, we selected three relatively common languages from MasakhaNER 2.0's expanded set: isiZulu (zul), Bambara (bam), and Setswana (tsn). For contrast, we also included two least common languages: Ghomala' (bbj) and Fon (fon)~\cite{Adelani2022MasakhaNER2A}. For each language, we begin with 50 samples selected to reflect the dataset's entity distribution and diversity. We devise a sampling strategy (cf. Supplementary Material) that ensures a representative benchmark for the evaluation of LLM performance, with variations in the number and type of entities.

For our annotation setting, we filtered the samples according to the proportions of the entities. We set a minimum threshold of 5~\% to ensure a sufficient presence of named entities for training. An upper threshold of 50~\% was chosen to focus the sampled subset on linguistically rich examples with meaningful entity interactions. Through dataset analysis, we found that samples with higher proportions of entities frequently consisted of simple lists, lacking contextual information crucial for robust model learning, leading to our choice of the upper threshold.

\begin{figure}[ht]
    \centering
    \begin{subfigure}[b]{0.45\textwidth}
        \includegraphics[width=\textwidth]{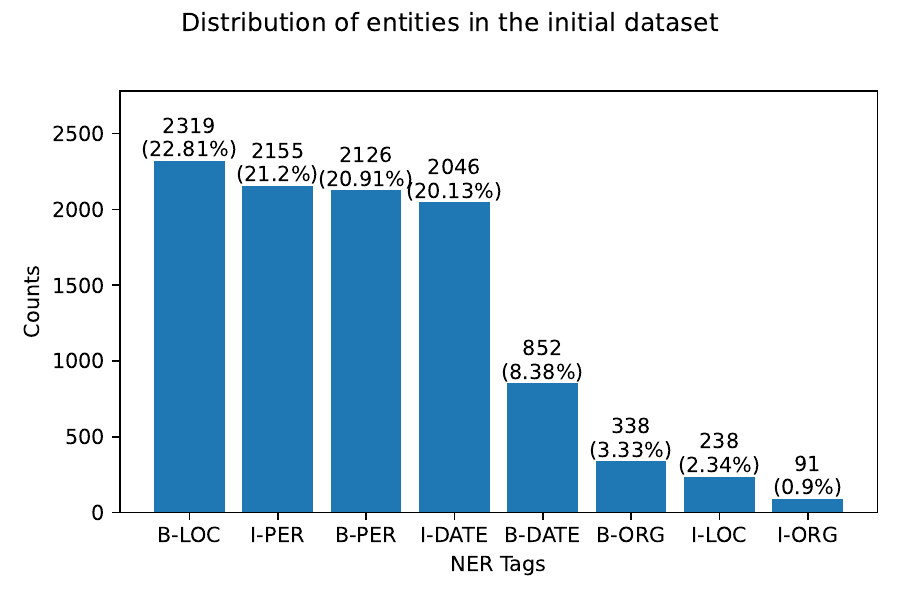}
        \caption{Initial entity distribution in the Bambara dataset.}
        \label{fig:sub1}
    \end{subfigure}
    \hfill
    \begin{subfigure}[b]{0.45\textwidth}
        \includegraphics[width=\textwidth]{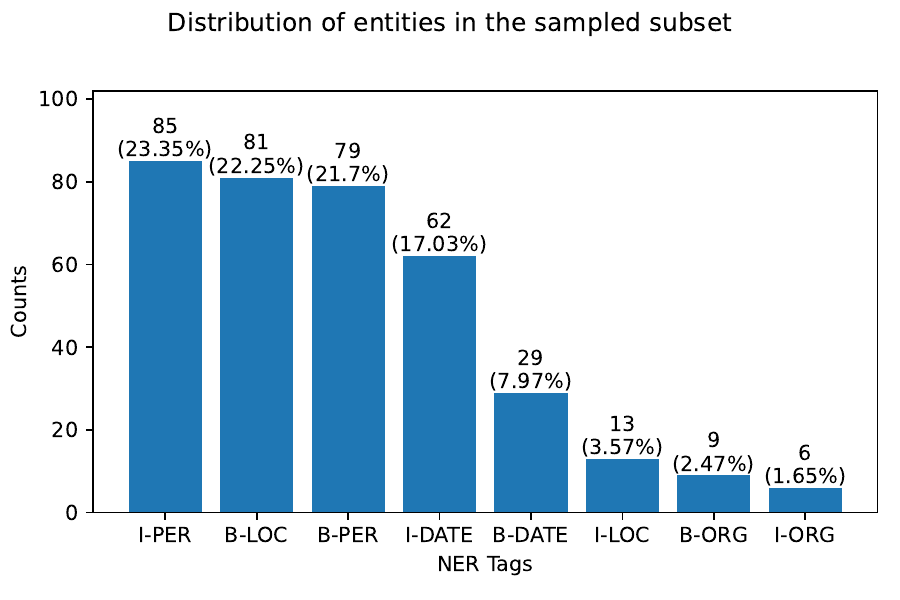}
        \caption{Entity distribution in 50 records sampled from Bambara dataset.}
        \label{fig:sub2}
    \end{subfigure}
    \caption{Entity distribution in a) the overall Bambara dataset and b) in 50 sampled records using (cf. Supplementary Material). Non-entities are excluded from the chart.}
    \label{fig:sampling_approach}
\end{figure}
Our balanced sampling approach closely preserves the entity distribution of the original dataset, enhancing the representativeness and reliability of subsequent results derived using this subset.

\subsubsection{Querying LLMs}
We employ a few-shot prompting approach~\cite{brown2020language} to extract entity labels from large language models. Our prompt template includes placeholders for the language, annotation examples, and sample tokens. 

The prompt directs the LLM to label each token, providing examples in the target language. It specifies the desired output format and set of labels. A zero-shot approach might not produce the desired format, leading to additional post-processing. Therefore, we opted for a few-shot approach, providing examples and instructions to ensure the model outputs match the required format. 

To ensure alignment with the MasakhaNER 2.0 annotation framework, we explicitly instruct the LLM to follow the MUC-6 annotation guidelines\footnote{\url{https://cs.nyu.edu/~grishman/muc6.html}}, the standard used by the original annotators of the dataset. These guidelines offer a consistent framework for recognizing and classifying named entities, such as persons, organizations, and locations.


In~\Cref{fig:prompt_template_short}, we present a condensed form of the few-shot prompt for NER annotation utilized in our research, and the full form of the prompt is shown in (cf. Supplementary Material).
\begin{figure}
\begin{mdframed}
\noindent Your task is to label entities in a given text written in \{language\}. Use the following labels for annotation: [...]

 Your task is to provide a list of named entity labels, where each label corresponds to a token. [...] in format [token, label].

\{examples\}

\noindent Note: [...]
\begin{itemize}[topsep=0pt,partopsep=0pt,parsep=0pt,itemsep=0pt]
    \item Follow MUC-6 (Message Understanding Conference-6) Named Entity Recognition (NER) annotation guidelines.
\end{itemize}
[...]
This is your sentence: \{sentence\}.
\end{mdframed}

\caption{Condensed version of the full prompt template with non-essential details abbreviated as [...].}
\label{fig:prompt_template_short}
\end{figure}

\subsubsection{Correct Output Format}
\label{subsubsec:correct_output_format}

Accurate data labeling in the correct format is crucial for automated use without manual intervention. To measure the proportion of correctly formatted labels, we repeat the annotation of the sampled records 10 times using multiple LLMs. We then compute labeling errors.

Two primary issues arise when querying LLMs for entity labels, namely, a) complete absence of a response and b) omission errors, where the model skips tokens that should be labeled. Omission errors also include the creation of extra tokens, leading to a mismatch between the number of predicted labels and tokens in the data sample.

As shown in \Cref{tab:missing_annotations_and_mismatched_tokens}, omission errors occur across all models and most languages in our experimental setup. GPT-4-Turbo performs best, with no more than 3 out of 50 samples per language exhibiting an omission error in at least one annotation attempt. Llama 2 70B, Starling-LM, and Mistral 7B demonstrate the poorest results, with significant mismatches between predicted labels and token counts. This indicates inconsistent output and difficulty in reliably following prompt instructions. The Gemini 1.0 pro model was unsuitable for this task due to its frequent inability to generate tokens by providing empty response, therefore preventing the assessment of omission errors.

\begin{table}[h]
    \centering
    \caption{Number of samples out of 50 with at least one omission error or missing prediction (output). Dashes (--) indicate that we could not calculate amount of omission errors because there was no response from the LLM.}
    \label{tab:missing_annotations_and_mismatched_tokens}
    \begin{tabular}{lccccc|ccccc}
        \toprule
        Model & \multicolumn{5}{c|}{Empty Predictions} & \multicolumn{5}{c}{Omission Error} \\
        \cmidrule(lr){2-6} \cmidrule(lr){7-11}
         & zul & bam & tsn & fon & bbj & zul & bam & tsn & fon & bbj \\
        \midrule
        GPT-4-Turbo & 0 & 0 & 0 & 0 & 0 & 0 & 3 & 1 & 0 & 2 \\
        Gemini 1.0 & 50 & 50 & 50 & 17 & 15 & -- & -- & -- & 4 & 1 \\
        Llama 2 70B & 0 & 2 & 2 & 1 & 0 & 12 & 15 & 18 & 38 & 13 \\
        Starling-LM & 0 & 2 & 0 & 1 & 0 & 10 & 24 & 5 & 15 & 7 \\
        Mistral 7B & 0 & 6 & 0 & 6 & 0 & 20 & 25 & 20 & 28 & 6 \\
        \bottomrule
    \end{tabular}
\end{table}

\Cref{tab:missing_annotations_and_mismatched_tokens} shows that while GPT-4-Turbo consistently provided annotations for all records in all annotation attempts, Gemini 1.0 failed to return annotations for all samples in isiZulu (zul), Bambara (bam) and Setwana (tsn) subsets. In this case, Gemini 1.0 was returning an empty response. 

Data samples with the aforementioned labeling issues are excluded from further analysis.

\subsubsection{Inter Annotator Agreement}
\label{sec:iaa}

As LLMs exhibit stochastic behavior when generating output~\cite{yuan2022selecting}, we measure the Inter-Annotator Agreement to assess the labeling consistency of these models. We proceed with the assumption that an effective model should label data consistently -- indicative of `no doubt' in its annotations -- thereby achieving a high IAA score. 

To evaluate IAA, the creators of the MasakhaNER 2.0 dataset utilized Fleiss' Kappa~\cite{fleisskappa} at the entity level~\cite{Adelani2022MasakhaNER2A}. Following this approach, we compare IAA between model-annotated and human-annotated data. Instead of relying on multiple human annotators, we instruct the model to reannotate the same samples 10 times. Additionally, we adjust the temperature setting to 0.1 to reduce randomness and limit creativity in the model's outputs.

\begin{table}
    \caption{Inter-annotator agreement (Fleiss' kappa) for model reannotations. IAA is measured for a representative subset of 50 samples and 10 reannotation attempts. For comparison, we also include IAA of human annotators from~\cite{Adelani2022MasakhaNER2A} for the whole dataset.}
    \label{tab:iaa}
    \centering
    \begin{tabular}{lccccc}
        \toprule
        Model & zul & bam & tsn & fon & bbj \\
        \midrule
        GPT-4-Turbo & \textbf{0.979} & 0.94 & 0.932 & 0.898 & 0.923 \\
        Gemini 1.0 & -- & -- & -- & 0.928 & 0.966 \\
        Llama 2 70B & 0.836 & 0.88 & 0.867 & 0.595 & 0.749 \\
        Starling-LM & 0.867 & 0.849 & 0.914 & 0.88 & 0.89 \\
        Mistral 7B & 0.907 & 0.877 & 0.922 & 0.83 & 0.906 \\
        Human annotation ~\cite{Adelani2022MasakhaNER2A} & 0.953 & \textbf{0.98} & \textbf{0.962} & \textbf{0.941} & \textbf{1.0} \\
        \bottomrule
    \end{tabular}
\end{table}

As illustrated in \Cref{tab:iaa}, GPT-4-Turbo and Gemini 1.0 demonstrate the highest IAA scores for the sampled subset of records, which are comparable to human IAA scores across the entire dataset. Specifically, in the isiZulu (zul) subset, GPT-4-Turbo achieved a higher IAA score compared to that of human annotators.
These results suggest that in some cases, LLMs are capable of reaching IAA scores comparable to, or even surpassing, those of human annotators.

\subsubsection{Consistency}
\label{sec:consistency}
To evaluate annotation quality and address model response variability, we introduce a metric to compute the consistency of prediction accuracy, referred to simply as consistency. This metric quantifies how frequently a model's predictions match the ground truth labels for each token over multiple reannotation attempts. Consistency, in this context, indicates the model's ability to produce accurate predictions reliably across multiple inferences on a single data sample. We posit that an effective model should demonstrate certainty and correctness in its labeling, reflected in a high consistency score.

We calculate the consistency score by comparing predicted labels against ground truth labels for each token.  This involves querying the same data sample multiple times and determining the proportion of instances where each token's predicted label matches its ground truth counterpart. We then average these per-sample scores to obtain the overall consistency for a data subset.

Cases with missing annotations or mismatches between predicted label count and data sample tokens are treated as complete mismatches for all ground truth labels. Thus, our consistency score also incorporates instances of omission errors.

\begin{table}
    \centering
    \caption{Consistency of prediction accuracy: it measures the model's ability to generate accurate annotations for a data sample across single or multiple inference attempts. It was evaluated using a representative subset of 50 records. Here, `10 iterations' refers to ten re-annotation attempts for a single data sample, while `1 iteration' indicates a single annotation attempt.}
    \label{tab:consistency}
    \begin{tabular}{lccccc}
        \toprule
        Model & zul & bam & tsn & fon & bbj \\
        \midrule
        \multicolumn{6}{c}{\textbf{10 iterations}} \\
        \midrule
        GPT-4-Turbo & \textbf{93.5} & \textbf{90.0} & \textbf{89.7} & \textbf{86.9} & \textbf{86.7} \\
        Gemini 1.0 & 0.0 & 0.0 & 0.0 & 61.2 & 62.6 \\
        Llama 2 70B & 17.9 & 18.1 & 14.7 & 15.8 & 55.5 \\
        Starling-LM & 66.6 & 54.1 & 63.6 & 70.9 & 76.6 \\
        Mistral 7B & 54.3 & 40.5 & 42.7 & 44.8 & 73.8 \\
        \midrule
        \multicolumn{6}{c}{\textbf{1 iteration}} \\
        \midrule
        GPT-4-Turbo & 93.48 & 91.35 & 89.7 & 86.89 & 85.91 \\
        Claude 3 Sonnet & 92.23 & 93.21 & 86.2 & 66.44 & 88.65 \\
        Claude 3 Opus & \textbf{97.86} & \textbf{94.64} & \textbf{92.87} & \textbf{95.1} & \textbf{89.2} \\
        \bottomrule
    \end{tabular}
\end{table}

As demonstrated in Table~\ref{tab:consistency}, the annotations produced by GPT-4-Turbo for 10 iterations exhibit significantly higher consistency compared to those obtained from other LLMs. Meanwhile, Llama 2 70B provides most inconsistent results, suggesting that this LLM may not be well suited for low-resource NER tasks.

Due to limited resources, we evaluate Anthropic's Claude 3 Sonnet and Claude 3 Opus using 50 samples and just one annotation attempt. As we calculate the number of times the label is predicted correctly, with only one annotation attempt, consistency is equivalent to accuracy.

\Cref{tab:consistency} demonstrates that Claude 3 Opus outperforms GPT-4-Turbo, achieving greater accuracy. However, due to significantly higher costs for Claude 3, we opt to continue with GPT-4-Turbo for our next experiments. 
Claude 3 Opus incurs a cost that is 1.5 times greater than that of GPT-4-Turbo for input tokens, and 2.5 times greater for output tokens.

\subsubsection{LLM Evaluation Results}

For subsequent experiments, GPT-4-Turbo was selected due to its consistently superior performance in our initial evaluations. Across these evaluations, it demonstrated the highest annotation accuracy, correct formatting, and the lowest rate of omission errors. Additionally, for the most popular languages within our test set, GPT-4-Turbo's performance was comparable to IAA scores achieved by human annotators.

While Claude 3 Opus outperformed GPT-4-Turbo on a smaller-scale benchmark, this needs to be assessed with more iterations for reliable validation.

Gemini 1.0 achieved higher IAA scores than GPT-4-Turbo for two less \nk{least} represented languages in the MasakhaNER 2.0 dataset \-- Ghomala' (bbj) and Fon (fon). However, Gemini 1.0's frequent empty responses warrant further investigation to understand potential limitations when handling specific annotation queries.

Llama 2 70B, Mistral 7B, and Starling-LM provided the least consistent results. The high number of omission errors with them suggests potential difficulty in handling complex annotation tasks. Interestingly, Starling-LM outperforms the other two models despite having fewer parameters.

\subsection{Effect of Prompt Design and Querying LLMs in batches}
\label{sec:querying_in_batches}

Before executing the experiments described in our methodology, we assess the potential impact that a carefully designed prompt template can have on the accuracy of annotations. In this section, we evaluate the consistency of annotations provided by Gemini 1.0 across a sample of 50 records, each subjected to 10 reannotation attempts. This evaluation contrasts our standard prompt template (cf. Supplementary Material), against a more concise version of the template that omits certain details. This comparison aims to discern how the level of detail in prompt templates affects annotation quality.
For the less concise prompt, we removed the detailed explanation of the output format but kept examples and the remaining parts intact. The following parts were removed from the prompt:
\begin{itemize}
    \item "You will receive a list of tokens as the value for the `input' key and text language as the value for the `language' key in a JSON dictionary."
    \item "The input tokens are provided in a list format and represent the text."
    \item "Your task begins now!"
\end{itemize}

As can be inferred from~\cref{tab:short_template}, providing additional details regarding the task, as well as the input and output formats, appears to enhance the accuracy of data annotation.

\begin{table}
    \centering
    \caption{Comparison of consistency for Gemini 1.0 using our default and short template.}
    \label{tab:short_template}
    \begin{tabular}{lccccc}
        \toprule
        Prompt type & bbj & swa \\
        \midrule
        Default prompt & 69.99 & 93.87 \\
        Shortened prompt  & 60.36 & 89.40 \\
        \bottomrule
    \end{tabular}
\end{table}

To minimize the usage of computational resources when annotating data using GPT-4-Turbo, we adopt a novel approach by querying the model in batches. Instead of processing just one sentence at a time, we combine two sentences in the same prompt and instruct the model to provide NER annotations for both samples simultaneously.

To evaluate this approach and compare the processing of individual records to batches of samples, we compute consistency scores using the methodology outlined in~\cref{sec: foundation_model_selection}. We evaluated two models: GPT-4-Turbo on isiZulu and Gemini 1.0 on a Swahili subset containing 50 records that are reannotated 10 times. As illustrated in~\cref{tab:batch_consistency}, batch annotation does not significantly reduce the consistency of results. Therefore, we opted to adopt this approach for further use.

\begin{table}
    \centering
    \caption{Consistency for annotating one sample at a time compared to a batch of samples.}
    \label{tab:batch_consistency}
    \begin{tabular}{lccccc}
        \toprule
        Method & zul - GPT-4-Turbo & swa - Gemini 1.0 \\
        \midrule
        One sample & 93.53 & 93.86 \\
        Batch of samples & 93.1 & 91.78  \\
        \bottomrule
    \end{tabular}
\end{table}

\subsection{Data Contamination}

To ensure the validity of the MasakhaNER 2.0 dataset for our experiments, we assessed the potential contamination of GPT-4-Turbo's training data with this dataset. We employ two methods:

a) \textit{Few-Shot In-Context Learning (ICL)}: Following Golchin et al. (2024) \cite{golchin2024time}, we use GPT-4 ICL to detect if the model can reproduce text from MasakhaNER 2.0.  This approach leverages the assumption that if an LLM's training data includes a specific dataset, the LLM can complete a partial text sample from that dataset with an exact or near-exact match. We also include CoNLL-2003 \cite{tjong-kim-sang-de-meulder-2003-introduction} and WikiANN \cite{rahimi-etal-2019-massively} datasets for comparison.

b) \textit{Contamination Score:}  To address the binary limitation of the ICL method, we introduce a `contamination score'. We randomly select 30 records across multiple languages from each dataset (details in the following paragraph). GPT-4-Turbo is then tasked with identifying the source dataset of each record. The corresponding prompt template is shown in~\cref{fig:prompt_identify_dataset}.

\begin{figure}
\begin{mdframed}
Identify the source (multilingual) NER dataset for this sample. Respond with the dataset name alone. \{sentence\}
\end{mdframed}
\caption{Prompt template for instructing the LLM to identify the source dataset of a given data sample. The placeholder \{sentence\} is replaced with the actual data sample for this task. For multilingual datasets, the term `multilingual' is specified in the prompt, while it is omitted for monolingual datasets.}
\label{fig:prompt_identify_dataset}
\end{figure}

The contamination score is the average proportion of correct identifications across three experimental runs.

For multilingual datasets (WikiANN, MasakhaNER 2.0), we sample 10 records each from three randomly selected languages (Bihari, Estonian, Slovak for WikiANN; isiXhosa, Ewe, Chichewa for MasakhaNER 2.0). 
We set the temperature parameter to 0 in GPT-4 and GPT-4-Turbo to minimize response randomness.

For the ICL method, we report~(cf.~\Cref{tab:data_contamination}) the number of exact or near-exact matches found by GPT-4 (as per Golchin \etal) and verified manually.

\begin{table}
    \centering
    \caption{Data Contamination Evaluation. We report the data contamination score and its standard deviation. The second and third rows display the number of exact and near-exact matches, respectively, evaluated as per  \cite{golchin2024time}. In case the human judgement about matches differs from that of GPT-4 judgement, the display format shows the GPT-4 judgment followed by the human judgment in parentheses.}
    \label{tab:data_contamination}
    \begin{tabular}{lccccc}
        \toprule
        Method & \makecell{CoNLL- \\ 2003} & \makecell{WikiANN \\ Multilingual} & \makecell{WikiANN \\ English} & \makecell{MasakhaNER \\ 2.0}  \\
        \midrule
        Contamination Score & $0.94\pm0.02$ & $0.81\pm0.13$ & $0.7\pm0.1$ & $0.04\pm0.02$  \\
        Exact matches \cite{golchin2024time}  & 0 & 0 & 3 & 0\\
        Near-exact matches \cite{golchin2024time}  & 2(3) & 0(1) & 1 & 0 \\
        \bottomrule
    \end{tabular}
\end{table}

Our findings suggest that GPT-4-Turbo training data might be contaminated with $4~\%\pm2~\%$ of the MasakhaNER 2.0 dataset, indicating minimal influence on the overall results, in contrast to the CoNLL-2003 and WikiANN datasets, which show significantly higher levels~(94~\% and 81~\% respectively) of potential contamination.

\subsection{Active Learning}

In this segment, we compare the performance of GPT-4-Turbo generated annotations against manual annotations within an active learning setting. Our baseline is a task-specific model trained on the entire manually annotated ground truth data. 

Due to significant class imbalance, our primary metric is entity-class accuracy, excluding non-entities where the model excels due to their overrepresentation. We do not report F1 computed at the entity level, where partial detection counts as a mismatch, as our preprocessing further splits entities into subtokens, complicating complete entity span detection. Accuracy, on the other hand, directly assesses the model's ability to identify and classify entities, even when an entity extends across multiple subtokens.

We establish a baseline accuracy score by training the model on the entire Bambara dataset (82~\% accuracy)~\cite{Adelani2022MasakhaNER2A}. We then fine-tune the model using active learning with ground truth annotations. In each iteration, we use entropy-based uncertainty sampling (cf.~\Cref{sec:method}) to select an additional 5~\% of the dataset. We repeat the active learning process but instead utilize annotations generated by GPT-4-Turbo, fine-tuning the task-specific model in each iteration.

\begin{figure}[h]
    \centering
    \includegraphics[width=\textwidth]{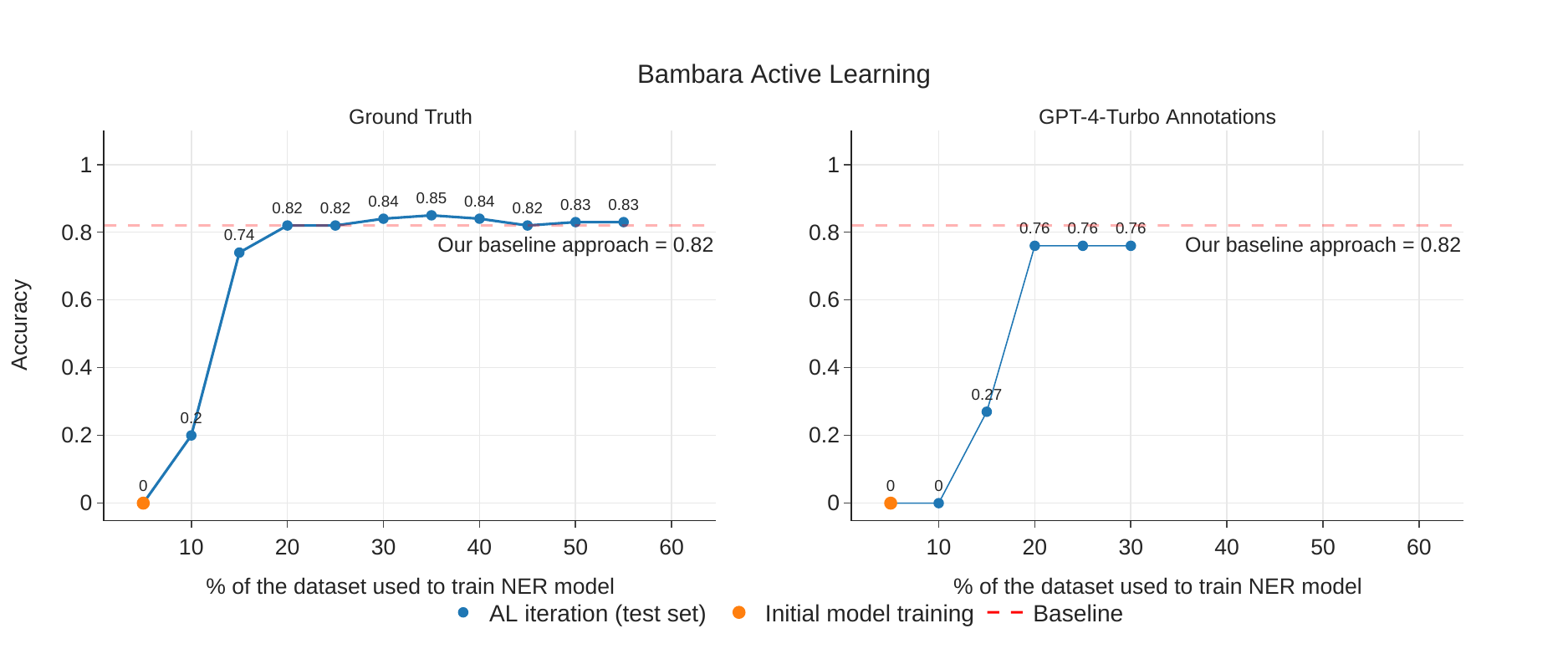}
    \caption{Accuracy (without non-entities) for Bambara test set achieved by using ground truth (left) and GPT-4-Turbo annotations (right) in our active learning framework. X-axis denotes the percentage of the dataset used for active learning iteration, and red dashed line represents our baseline - simple AfroXLMR-mini training using 100~\% of the dataset without active learning.}
    \label{fig:bam-al}
\end{figure}

Our active learning approach with ground truth annotations reaches the baseline performance after utilizing only 20~\% of the data and surpasses it after 30~\%. In contrast, using GPT-4-Turbo annotations, we achieve 76~\% accuracy but don't fully reach the baseline within 5 iterations (cf.~\Cref{fig:bam-al}). This difference may be due to the lower average accuracy of GPT-4-Turbo annotations (84.5(9)~\% compared to manual ground truth). We observe similar trends for isiZulu (cf. Supplementary Material). Active learning with manual ground truth approaches the baseline, while using GPT-4-Turbo annotations with 30~\% of the data falls relatively short. 
It is to be noted that since the baseline here is with human annotations, the ground truth will always be more numerically accurate, and there will always be a disparity when comparing annotations from other sources. 

To assess the cost comparison between human annotation and GPT-4-Turbo annotation for the Bambara and isiZulu train datasets, we based our estimates on the price per 200 annotated sentences as detailed in~\cite{Adelani2022MasakhaNER2A}. The estimated cost for human annotation for Bambara train set is USD 233.5, and for isiZulu train set is USD 292.5 per annotator. These costs are then multiplied by the number of annotators as indicated in the study.

Annotating 30~\% of the MasakhaNER 2.0 dataset with GPT-4-Turbo, in comparison, is significantly cheaper than human annotation, costing approximately USD 16.5 per language. This makes our approach approximately 42.45 times less expensive for Bambara and 53.18 times less expensive for isiZulu. 

Our approach demonstrates the potential of GPT-4-Turbo for cost-effective dataset annotation. Although there is an accuracy trade-off compared to ground truth, the substantial cost reduction makes it a compelling option, especially in resource-constrained settings. GPT-4-Turbo's performance is promising, and with continued training, we expect the accuracy and language coverage of state-of-the-art LLMs to improve further, making automated data labelling an increasingly viable alternative.

\section{Conclusion}

In this work, we analyzed foundation models in the context of low-resource NER data annotation. Our findings highlight the intricate balance of trade-offs and potential benefits associated with using LLMs for this task. Evaluating various LLMs revealed performance variations, with GPT-4-Turbo demonstrating consistent and accurate annotations, while others showed strengths in specific contexts. This emphasizes the necessity of selecting appropriate models based on task requirements.

Common challenges across LLMs were identified, including token skipping and the need for refinement in generating correctly formatted outputs. Addressing these challenges through further training or advanced prompt engineering could enhance model reliability. 

Although models like GPT-4-Turbo may exhibit slight accuracy reductions compared to human annotations, it is crucial to note that the baseline for comparison here is human annotations themselves. Consequently, human annotations are inherently expected to be numerically superior in accuracy. Additionally, human annotations are not infallible due to subjectivity and potentially outdated annotation guidelines. This makes minor accuracy reductions from LLM annotations less significant, especially considering the substantial cost savings they offer.

Demonstrating GPT-4-Turbo's cost-effectiveness for data annotation, our study indicates potential cost savings of at least 42.45 times compared to human annotation. This is particularly compelling in resource-constrained settings, where minor accuracy trade-offs may be acceptable given the significant benefits. Furthermore, GPT-4-Turbo's promising performance and the ongoing advancement of LLMs suggest even greater potential for their cost-effective integration in the future.

Furthermore, in this work, we introduced a novel methodology to assess data leakage and quantify the potential for data contamination in LLM training datasets. This is to assess whether the results from our experiments could be extrapolated to similar unseen low-resource language domains.



In conclusion, our investigation into the use of foundation models for data annotation in low-resource African languages reveals the transformative potential of LLMs in NLP. Despite the challenges, our study demonstrates that LLMs, particularly through active learning approaches, offer a cost-effective and efficient solution for NER tasks. 

As a future work, exploring next-generation LLMs may yield noticeable improvements, as Claude 3 Opus outperformed GPT-4-Turbo in our smaller-scale benchmark. Additionally, annotation can be done by multiple different LLMs, with labels assigned based on majority voting. This approach utilizes knowledge of different models, potentially reducing bias and errors that arise when using a single model.


\begin{credits}
\subsubsection{\ackname} This work has been supported by the OpenWebSearch.eu project, which is funded by the  EU under the GA 101070014. We thank the EU for their support. We also want to thank the
anonymous reviewers for their valuable feedback and comments.

\subsubsection{\discintname}
The authors declare no conflict of interest.
\end{credits}




\newpage
\appendix


\section{Algorithm for Representative Data Sampling}
\label{sec:appendix_sampling_algorithm}

We use~\Cref{alg:entity_sampling} to sample records from the MasakhaNER 2.0 dataset for evaluating the quality of LLM annotation. This approach ensures that the selected samples reflect the original dataset's entity distribution, while maintaining sufficient named entity count and avoiding overly simplistic lists of entities. It provides a reliable subset of data for assessing LLM performance across different languages.

\begin{algorithm}[H]
    \caption{Entity-based Balanced Sampling Approach}
    \label{alg:entity_sampling}
    \SetKwFunction{entitysampling}{EntitySampling}
    \SetKwProg{Procedure}{Procedure}{}{}
    \Procedure{\entitysampling{$D$: dataset}}{
        $D' \gets \emptyset$ \tcp*[r]{\textcolor{gray}{Initialize empty set}}
        \ForEach{$s \in D$}{
            $entities \gets \text{NER}(s)$ \tcp*[r]{\textcolor{gray}{Count named entities}}
            \If{$entities \neq \emptyset$}{
                $p \gets \frac{|entities|}{|s|}$\;\\
                \If{$0.05 \leq p \leq 0.50$}{
                    $D' \gets D' \cup \{s\}$\;
                }
            }
        }
        $\text{SortDescending}(D')$ \tcp*[r]{\textcolor{gray}{Sort by entity count}}
        $numRecords \gets |D'|$ \tcp*[r]{\textcolor{gray}{Total number of records in the dataset}}
        $\mu \gets \frac{numRecords}{2}$ \tcp*[r]{\textcolor{gray}{Mean of the normal distribution}}
        $\sigma \gets \frac{numRecords}{4}$ \tcp*[r]{\textcolor{gray}{Standard deviation of the normal distribution}}
        $w \gets []$ \tcp*[r]{\textcolor{gray}{Initialize weight array}}
        \ForEach{$s_i \in D'$}{
            $x_i \gets \text{index of } s_i \text{ in } D'$\;\\
            $w_i \gets \exp\left(-\frac{(x_i - \mu)^2}{2 \cdot \sigma^2}\right)$\; \tcp*[r]{\textcolor{gray}{Weight for $s_i$ using $\mathcal{N}(\mu, \sigma)$}}
            $w \gets w \cup \{w_i\}$ \tcp*[r]{\textcolor{gray}{Append weight to the array}}
        }
        $w \gets \text{Normalize}(w)$\; \tcp*[r]{\textcolor{gray}{Normalize weights}}
        $D_{\text{balanced}} \gets \text{WeightedSample}(D', w)$\;\\
        \Return $D_{\text{balanced}}$\;
    }
\end{algorithm}

\section{Prompt Template for Querying Foundation Model}
\label{sec:appendix_prompt_template}

To query LLMs we utilize a flexible prompt format, tailored for each language in MasakhaNER2.0 dataset. 

\Cref{fig:prompt_template} illustrates the prompt template used to query the foundation models for NER labels. The prompt template includes descriptions of the annotation labels, details of the required output format, and corresponding examples.

\begin{figure}
\begin{mdframed}
\noindent Your task is to label entities in a given text written in \{language\}. Use the following labels for annotation:
\begin{itemize}[topsep=0pt,partopsep=0pt,parsep=0pt,itemsep=0pt]
    \item "O": Represents words that are not part of any named entity.
    \item "B-PER": Indicates the beginning of a person's name.
    \item "I-PER": Used for tokens inside a person's name. 
    \item "B-ORG": Marks the beginning of an organization's name.
    \item "I-ORG": Tokens inside an organization's name.
    \item "B-LOC": Marks the beginning of a location (place) name.
    \item "I-LOC": Tokens inside a location name.
    \item "B-DATE": Marks the beginning of a date entity.
    \item "I-DATE": Tokens inside a date entity.
    \end{itemize}
    \vspace{4pt}
You will receive a list of tokens as the value for the 'input' key and text language as the value for the 'language' key in a JSON dictionary. Your task is to provide a list of named entity labels, where each label corresponds to a token. Output the tokens with their corresponding named entity labels in a JSON format, using the key 'output'. 'output' should contain a list of tokens and their entity labels in format [token, label].

\vspace{4pt}
\{examples\}
\vspace{4pt}

\noindent Note:
\begin{itemize}[topsep=0pt,partopsep=0pt,parsep=0pt,itemsep=0pt]
    \item The input tokens are provided in a list format and represent the text.
    \item Important: the output should be a list with the same length as the input list, where each element corresponds to the named entity label for the corresponding token. Do not change the order of tokens and do not skip them.
    \item The named entity labels are case-sensitive, so please provide them exactly as specified ("B-PER", "I-LOC", etc.).
    \item Follow MUC-6 (Message Understanding Conference-6) Named Entity Recognition (NER) annotation guidelines.
\end{itemize}
Your task begins now! \\
- Output JSON only. Enclose all tokens and tags in double brackets.\\
This is your sentence: \{sentence\}.
\end{mdframed}

\caption{Prompt template for querying foundation model.}
\label{fig:prompt_template}
\end{figure}

The prompt includes placeholders (\{language\}, \{examples\}, \{sentence\}) which are substituted with the specific language name, examples, and the data sample for annotation, respectively. 

\section{Annotation Examples for Querying Foundation Model}

\begin{figure}

\centering
\includegraphics[width=\textwidth]{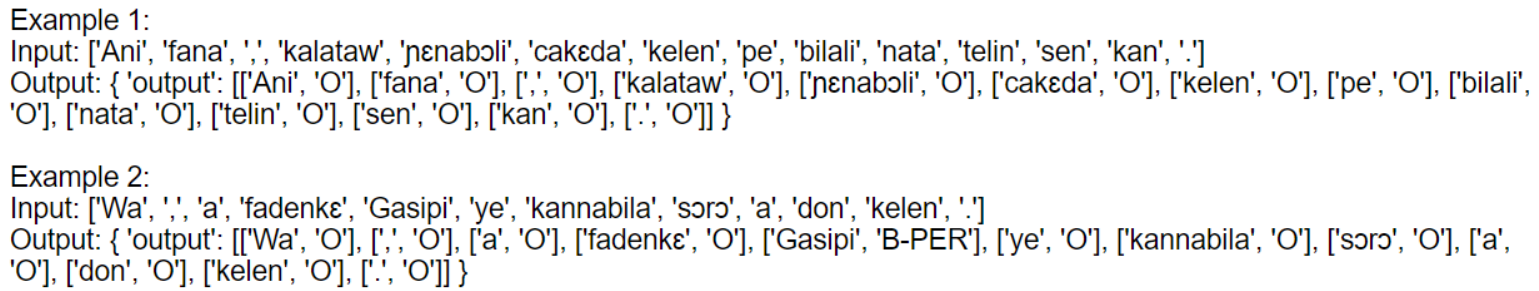}


\caption{Annotation examples for querying foundation model.}
\label{fig:ann_examples}
\end{figure}

For each language featured in MasakhaNER2.0, we generate specific annotation examples. Two records are randomly chosen from the MasakhaNER2.0 \cite{Adelani2022MasakhaNER2A} test set for each language: the first record contains no named entities, while the second includes at least one named entity. These examples are then structured as a single string for incorporation in place of the \{examples\} placeholder. \Cref{fig:ann_examples} shows an example of annotation for the Bambara language.

\section{Active Learning for iziZulu language}
\label{sec:appendix_active_learning_izizulu}

To validate the usability of our approach for different languages, we repeat the experiments with active learning using LLM annotations in the loop for the isiZulu language, as previously performed for Bambara. We establish a baseline accuracy score by training the model on the entire isiZulu dataset, achieving accuracy of 90\%.

We repeat the active learning process for isiZulu by fine-tuning the model first using ground truth annotations in the first experiment, and then using annotations generated by GPT-4-Turbo. Our active learning approach with ground truth annotations reaches close-to-baseline performance after utilizing 20\% of the data, and a maximum accuracy of 89\% is achieved using 35\% of the data.

The model trained using GPT-4-Turbo annotations on 30\% of the dataset falls relatively short.

\begin{figure}[h]
    \centering
    \includegraphics[width=\textwidth]{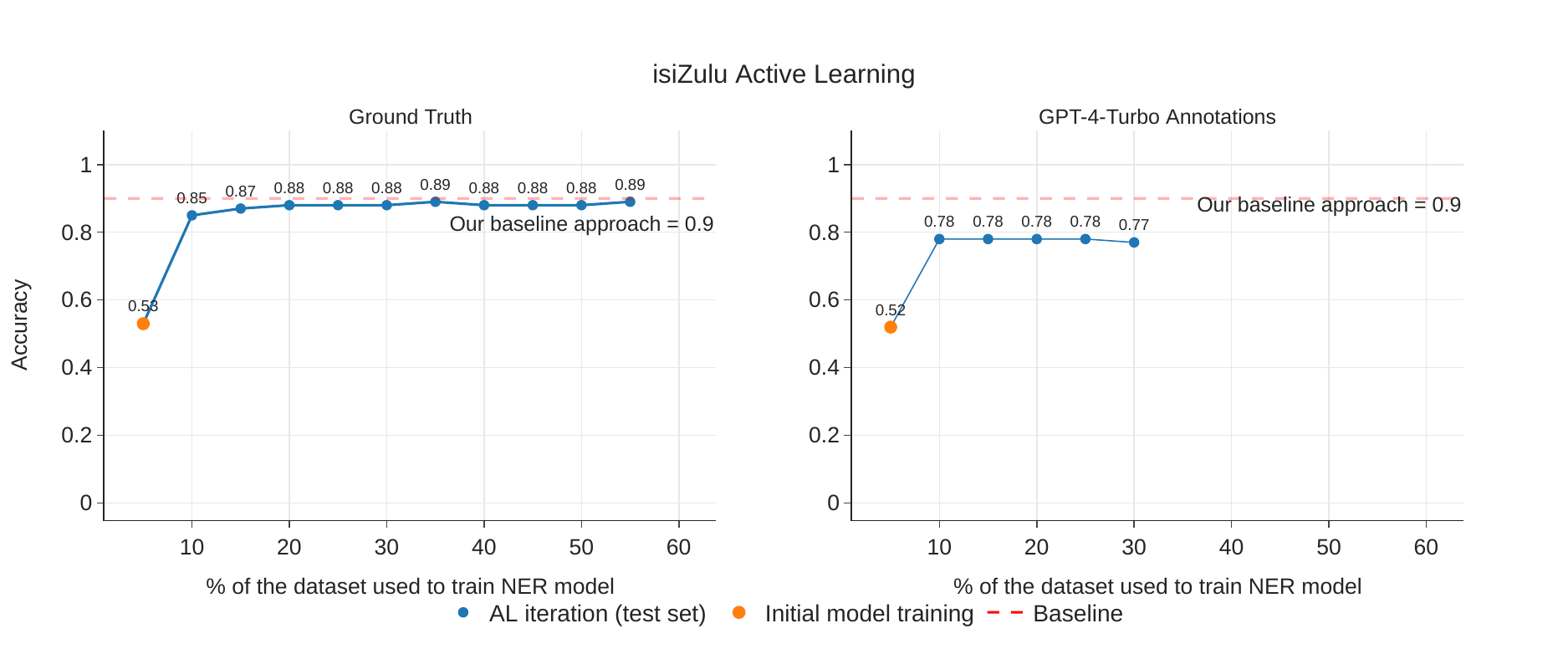}
    \caption{Accuracy (without non-entities) for isiZulu test set achieved by using ground truth (left) and GPT-4-Turbo annotations (right) in our active learning framework. X-axis denotes the percentage of the dataset used for active learning iteration, and red dashed line represents our baseline - simple AfroXLMR-mini training using 100~\% of the dataset without active learning.}
    \label{fig:zul-al}
\end{figure}


\end{document}